\documentclass[10pt,twocolumn,letterpaper]{article}
\usepackage{multirow}
\usepackage{cuted}
\usepackage{capt-of}
\usepackage{wacv}              
\usepackage[accsupp]{axessibility}  

\usepackage{graphicx} 
\usepackage{makecell}
\graphicspath{ {./figures/} } 

\newcommand{\Eref}[1]{Eq.~(\ref{#1})} 
\newcommand{\Fref}[1]{Fig.~\ref{#1}} 
\newcommand{\Tref}[1]{Table~\ref{#1}} 
\newcommand{\RNum}[1]{\lowercase\expandafter{\romannumeral #1\relax}}

\usepackage{pifont} 
\newcommand{\cmark}{\ding{51}} 
\newcommand{\xmark}{\ding{55}} 

\definecolor{wacvblue}{rgb}{0.21,0.49,0.74}
\usepackage[pagebackref,breaklinks,colorlinks,allcolors=wacvblue]{hyperref}

\def\wacvPaperID{2309} 
\def\confName{WACV}
\def\confYear{2026}

\title{Not Like Transformers: Drop the Beat Representation for Dance Generation with Mamba-Based Diffusion Model}

\author{
Sangjune Park$^{1}$ \quad
Inhyeok Choi$^{1}$ \quad
Donghyeon Soon$^{2}$ \quad
Youngwoo Jeon$^{1}$ \quad
Kyungdon Joo$^{1}$\thanks{\small{Corresponding author.}} \\
$^{1}$Ulsan National Institute of Science and Technology, South Korea \\
$^{2}$Daegu Gyeongbuk Institute of Science and Technology, South Korea \\
{\tt\small \{psj9116, inhyeok.choi, youngwoo.jeon, kyungdon\}@unist.ac.kr} \quad
{\tt\small dhsoon@dgist.ac.kr}
}

\begin{document}
\maketitle

\begin{abstract}
Dance is a form of human motion characterized by emotional expression and communication, playing a role in various fields such as music, virtual reality, and content creation.
Existing methods for dance generation often fail to adequately capture the inherently sequential, rhythmical, and music-synchronized characteristics of dance.
In this paper, we propose \emph{MambaDance}, a new dance generation approach that leverages a Mamba-based diffusion model.
Mamba, well-suited to handling long and autoregressive sequences, is integrated into our two-stage diffusion architecture, substituting off-the-shelf Transformer.
Additionally, considering the critical role of musical beats in dance choreography, we propose a Gaussian-based beat representation to explicitly guide the decoding of dance sequences.
Experiments on AIST++ and FineDance datasets for each sequence length show that our proposed method effectively generates plausible dance movements while reflecting essential characteristics, consistently from short to long dances, compared to the previous methods.
Additional qualitative results and demo videos are available at \small{\url{https://vision3d-lab.github.io/mambadance}}.
\end{abstract}

\section{Introduction} \label{sec:1}

\begin{figure}[ht]
    \centering
    \includegraphics[width=.95\linewidth]{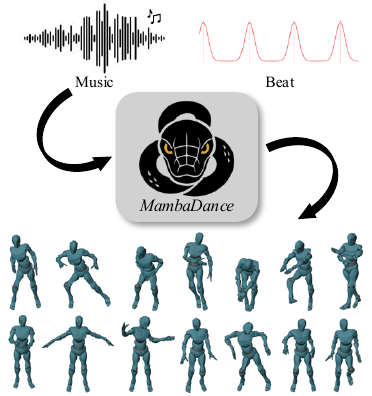}
    \caption{We propose \emph{MambaDance}, a Mamba-based two-stage diffusion framework with an informative Gaussian beat representation. The result is coherent, beat-synchronized motion across variable lengths on AIST++~\cite{fact} and FineDance~\cite{finedance}.}
    \label{fig:teaser}
\end{figure}

Generating expressive and realistic dance movements synchronized with music is a long-standing challenge. 
Dance serves as a medium of embodied communication and artistic expression, characterized by its structured, rhythmic nature. 
Traditionally, choreography requires manual effort from experts or frame-by-frame animation, both of which are costly and time-consuming.
Recent advances in generative models have enabled the automatic generation of music-driven dance~\cite{fact, dancetransformer, bailando, bailandopp, edge, lodge, popdg} with applications spanning entertainment, content creation, gaming, and virtual reality. 
These developments offer scalable alternatives to manual choreography, but also introduce technical challenges in modeling the temporal complexity and rhythmic sensitivity inherent in dance.

Transformer-based architectures~\cite{dancetransformer, fact, bailando, bailandopp, edge, lodge, popdg} are widely adopted in current 3D dance generation methods because of their ability to model global temporal dependencies. 
However, dance sequences require more than a general understanding of context.
They demand strong temporal causality and consistent autoregressive structure over extended time horizons.
Transformer~\cite{transformer} models fundamentally lack an inductive bias for sequential progression and often show inefficiency and inconsistency when generating long motion sequences. 
These limitations make it challenging to produce motion that is both rhythmically aligned and temporally coherent.

In addition to architectural limitations, the representation of musical beats plays a critical role in structuring dance motion.
For example, in instruction, choreography is first segmented into beat-length phrases (e.g., 8-counts), and learners practice following those anchors before performing to the full track with music.
Many existing methods~\cite{fact, bailando, edge, bailandopp, lodge, popdg} incorporate simple 1-dimensional beat features within the music feature vector.
While such representations provide coarse alignment, they do not explicitly model how beats influence the motion sequence. 
Beat-It~\cite{beatit} decouple beat cues from audio features to strengthen beat-dance alignment while introducing a beat-distance estimator with an auxiliary alignment loss.
Nevertheless, this approach encodes beat influence implicitly in network activations, rather than modeling an explicit temporal prior that shapes motion over time.
%
%
These two challenges, namely long and autoregressive sequence modeling and effective beat conditioning, may seem independent, but they are closely related. 
Inadequate sequence modeling weakens rhythmic alignment, and insufficient beat representations limit the ability of the model to produce structured and expressive movements. 
A unified approach is necessary to address both aspects simultaneously, enabling the generation of dance that is both temporally consistent and rhythmically synchronized.

As illustrated in \Fref{fig:teaser}, we propose \emph{MambaDance}, a music-to-dance generation framework built on a Mamba-based, two-stage diffusion architecture.
Unlike prior Transformer or hybrid designs, our model replaces attention entirely with state space modules to efficiently capture long and autoregressive dynamics. 
We follow the overall two-stage paradigm of Lodge~\cite{lodge}, but the decoder block in a denoising network of diffusion comprises (\lowercase\expandafter{\romannumeral1}) Single-Modal Mamba that processes motion features and (\lowercase\expandafter{\romannumeral2}) Cross-Modal Mamba that fuses motion with music, followed by Adaptive Linear Modulation.
%
To enforce rhythmic structure, we introduce a novel beat representation based on Gaussian decay, which provides smooth and interpretable temporal priors centered around musical beats. 
This representation enables the model to generate motion that follows the underlying rhythmic phrasing of the music.
We evaluate on AIST++~\cite{fact} and FineDance~\cite{finedance} with different sequence lengths to systematically validate superior performance across multiple sequence lengths, demonstrating length-robust synthesis.
Experimental results show that \emph{MambaDance} consistently generates dance motions that are more physically plausible and rhythmically aligned than the off-the-shelf methods.

In summary, our contributions are as follows:
\begin{itemize}
    \item We propose a novel framework, \emph{MambaDance}, a fully Mamba-based diffusion model designed for autoregressive music-to-dance generation.
    \item We present a Gaussian-based beat representation that explicitly encodes rhythmic structure to guide motion decoding, considering key properties of music beats.
    \item We conduct comprehensive experiments and ablation studies demonstrating that our approach consistently outperforms previous baselines across fidelity and rhythm alignment metrics.
\end{itemize}

\section{Related Work}
\label{sec:2}

\begin{figure*}[ht]
    \centering
    \includegraphics[width=0.95\textwidth]{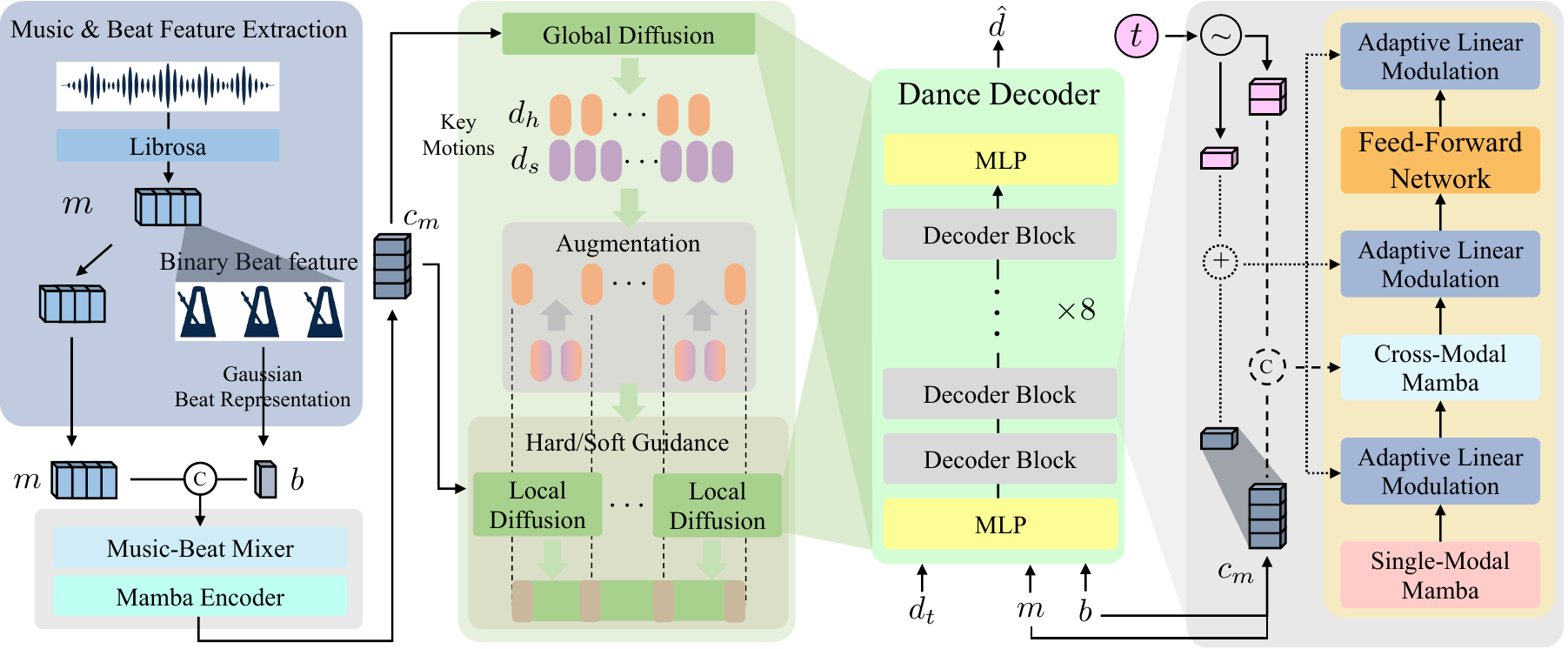}
    \caption{\textbf{The overall architecture of \emph{MambaDance}.} We extract music feature $m$, and a novel beat representation $b$ from the binary mask of beat of the feature (blue box). Two-stage diffusion architecture makes our approach enable length-agnostic generation in a single inference (green box). Decoder of the diffusion consists of the proposed Mamba~\cite{mamba, mamba2}-based modules, e.g., Single-Modal Mamba (SMM), Cross-Modal Mamba (CMM), and Adaptive Linear Modulation (AdaLM) (gray box).}
    \label{fig:overall_architecture}
\end{figure*}

\subsection{Human Motion Generation}

Human motion generation is crucial for realistic animation and interactive systems, drawing interest from vision, graphics, and robotics communities.
Earlier methods focused on motion retrieval and interpolation, yielding plausible but limited results for simple actions like walking.

Recent advances in deep generative models have enabled neural networks to generate more flexible and expressive human motion.
GAN-based methods~\cite{hp_gan} use adversarial training with multi-scale discriminators to enhance realism and model transition dynamics. Task-specific losses further help capture motion uncertainty.
Autoencoder-based approaches compress motion into latent spaces, often using recurrent or transformer models~\cite{actor, most} to align motion with language. Recent work improves expressiveness via body-part-aware vector quantization~\cite{motionclip,temos}.
Diffusion models have emerged as a strong alternative, offering realism and controllability. Text-conditioned models~\cite{motiondiffuse,mdm} treat motion as a denoising process, while latent-space variants~\cite{motion_latent_diffusion} improve efficiency and structure.
%
%

However, efficiently handling long, autoregressive sequences remains challenging, especially for transformer-based diffusion models due to scalability issues. Recent work such as MotionMamba~\cite{motionmamba} addresses this with a state-space model (SSM) that uses hierarchical temporal and bidirectional spatial modeling, improving both long-term consistency and generation speed.
Nevertheless, effectively modeling long sequences and incorporating temporal structure or rhythmic cues remain important challenges to be addressed.
\subsection{Dance Generation}

Early studies on music-driven dance generation regarded the task as an autoregressive modeling problem and adopted conventional machine learning techniques. 
While these approaches captured basic motion-music alignment, they lacked adaptability to diverse musical inputs and struggled to generate long and varied motion sequences.

Transformer-based models, such as FACT~\cite{fact}, have demonstrated strong capabilities in modeling temporal dynamics. In addition, approaches that combine VQ-VAE with GPT, exemplified by Bailando~\cite{bailando}, as well as diffusion-based frameworks including EDGE~\cite{edge}, Lodge~\cite{lodge}, and POPDG~\cite{popdg}, further enhance the realism, diversity, and editability of generated dance motions.
%
However, as these methods are based on Transformer architectures, they face challenges in modeling dance data, which is typically long and exhibits autoregressive dependencies.
%
Recently, several works~\cite{megadance, matchdance} adopt Mamba structure~\cite{mamba, mamba2} to well capture the autoregressive nature of dance data. However, most of these studies adopt hybrid architectures, where Mamba is only partially combined with Transformers. While such designs improve local continuity, they still inherit the quadratic complexity and discontinuities of attention.

\subsection{State-Space Model for Motion and Dance}







Recent approaches in dance generation have emphasized the importance of long-sequence modeling and synchronization with music. Transformer-based models have been widely adopted due to their ability to capture global dependencies, but they often struggle with autoregressive consistency and computational efficiency when generating long sequences.
State-space models (SSMs) have recently emerged as a promising alternative for sequence modeling. In particular, Mamba~\cite{mamba, mamba2} has been developed to efficiently model long-range autoregressive dependencies in sequential data such as human motion and dance. By leveraging time-varying parameters, Mamba captures long-term dependencies more effectively than Transformers while maintaining linear-time complexity. In addition, its structure inherently embeds a sequential inductive bias, which facilitates smooth temporal continuity.

Several recent works leverage Mamba structure for human motion and dance generation.
MotionMamba~\cite{motionmamba} utilizes hierarchical SSMs to enhance temporal coherence and spatial motion modeling in human motion generation. AlignYourRhythm~\cite{alignyourrhythm} integrates a rhythm-aware module based on Mamba to improve music-motion alignment. MegaDance~\cite{megadance} and MatchDance~\cite{matchdance} adopt Mamba-Transformer hybrid architectures, leveraging Mamba for local dependency modeling and Transformer to capture global context.
While these studies successfully exploit the advantages of Mamba, they adopt hybrid architectures, partially combining Mamba with Transformers.

In contrast, to the best of our knowledge, our work is the first to fully replace Transformer-based modules with Mamba-based modules for dance generation, including single-modal and cross-modal components.
We achieve structured, rhythmically aligned 3D dance generation with temporal consistency and improved computational efficiency, while fully leveraging autoregressive inductive bias of Mamba for long-sequence modeling.

\section{Method} \label{sec:method}
In this work, we propose \emph{MambaDance}, a two-stage diffusion framework that leverages effective long and autoregressive sequence modeling capacity of Mamba~\cite{mamba, mamba2} for music-driven 3D dance generation.
The model generates plausible and natural dance motion while exploiting the inductive bias to model long and autoregressive 3D motion sequences of the state-space model.
Considering the role of music beats that structure and anchor movements in choreography, we propose a new beat representation that explicitly enhances beat influence.
This representation provides an intuitive control signal and improves rhythmic guidance during dance generation.

As illustrated in \Fref{fig:overall_architecture}, \emph{MambaDance} takes music and beat features extracted both from single raw music, and outputs 3D dance sequences.
To produce a long sequence of dance with a single inference, a global diffusion generates key motions.
The key motions are augmented by mirroring and connecting adjacent motions.
Then a local diffusion generates detailed movements based on the primitives with hard/soft guidance.
The two diffusion models share same dance decoder structure, which consists of two MLPs and 8 decoder blocks.
Each decoder block mainly comprises (\RNum{1}) a Single-Modal Mamba (SMM), (\RNum{2}) a Cross-Modal Mamba (CMM), and (\RNum{3}) Adaptive Linear Modulation (AdaLM), which is fully designed with Mamba while substituting all attention modules.

\subsection{Mamba for Dance Generation} \label{sec:mamba_for_dance_generation}
Following prior works~\cite{edge, lodge}, we represent sliced dance motion as a sequence of SMPL~\cite{smpl, smplx} poses $d \in \mathbb{R}^{l \times D_\text{motion}}$.
Here, $l \in \{N, n\}$ denotes length sliced from total $L$-length motion or music sequences, where $N$ and $n$ for global and local diffusion, respectively.
Given music feature $m \in \mathbb{R}^{l \times D_\text{music}}$ and beat representation $b \in \mathbb{R}^{l \times 1}$, an MLP-based Music-Beat Mixer and a Mamba encoder produce a beat-highlighted condition $c_m \in \mathbb{R}^{l \times E}$ that injects other modality information during decoding.
In a Mamba-based dance decoder, an input MLP lifts a noisy motion sequence $d_t$ in timestep $t$ to a latent $z_d \in \mathbb{R}^{l \times E}$, a stack of decoder blocks processes the latent with conditions, and an output MLP projects back to $\hat{d} \in \mathbb{R}^{l \times D_\text{motion}}$.
Each decoder block contains a \emph{Single-Modal Mamba (SMM)}, a \emph{Cross-Modal Mamba (CMM)}, a lightweight feed-forward network (FFN), and \emph{Adaptive Linear Modulation (AdaLM)}.
Replacing self-attention with SSM and cross-attention with CMM yields linear-time sequence processing and stable long-horizon generation compared with attention-based designs~\cite{mamba, mamba2}.

\begin{figure}
    \centering
    \includegraphics[width=0.46\textwidth]{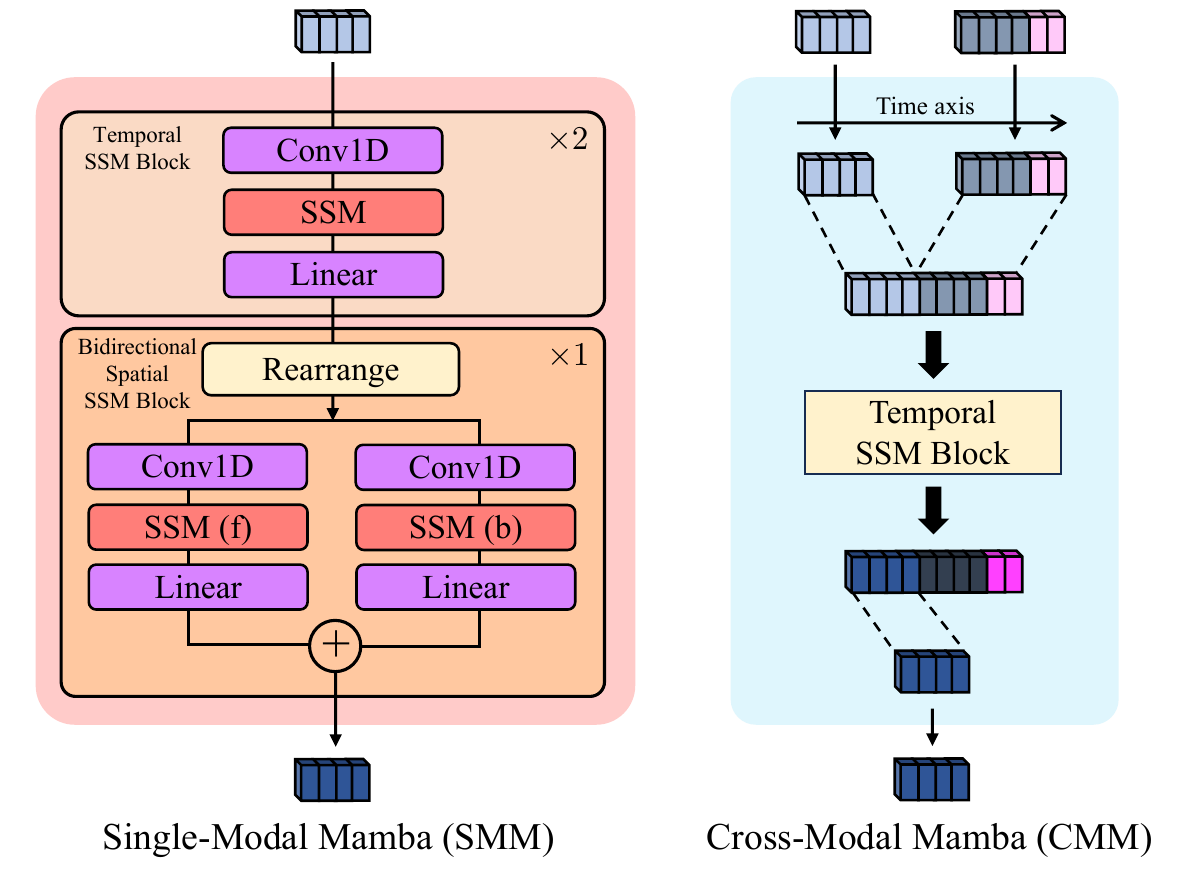}
    \caption{\textbf{Single-Modal Mamba (left) and Cross-Modal Mamba (right).} For the input sequences to the Cross-Modal Mamba, Light blue, dark blue, and pink blocks correspond to motion, condition, and timestep tokens, respectively.}
    \label{fig:smm_cmm}
\end{figure}
Single-Modal Mamba (SMM) operates solely on motion latents, whereas Cross-Modal Mamba (CMM) fuses motion latents with musical and diffusion timestep tokens (\Fref{fig:smm_cmm}).
SMM transforms input noisy motion latent with two Temporal SSM Blocks and a Bidirectional Spatial SSM Block.
The Temporal SSM Block, following ~\cite{mamba, mamba2}, propagates a sequence $a \in \mathbb{R}^{l \times E}$ along the length axis $l$.
The Spatial SSM Block rearranges a sequence to $a' \in \mathbb{R}^{E \times l}$ and propagates along the latent-channel axis $E$ in a bi-directional manner to encourage cross-channel coordination.
In contrast, CMM performs cross-modal integration by concatenating the motion latent $z_d$, the musical condition $c_m$ (mixture of music and beat information), and the diffusion timestep tokens $e_t \in \mathbb{R}^{2 \times E}$ as $[z_d, c_m, e_t]$, applying a Temporal SSM over the concatenated sequence, and slicing the motion part of the output.
These modules serve as an effective alternative to attention modules, especially for the 3D dance data:long, complex, and time-sequential.

We introduce Adaptive Linear Modulation (AdaLM), an alternative to FiLM~\cite{film}, which is a simple normalization-based modulator that conditions 1D input sequences.
Analogous to adaptive normalization techniques~\cite{adain, guided_diffusion} in the 2D vision, we apply feature-wise affine modulation to group-normalized latents:
\begin{equation} \label{eq:linear_mod}
\begin{split}
    \operatorname{AdaLM}(z_d, \gamma, \beta) = (1 + \gamma) \operatorname{GroupNorm}(z_d) + \beta,
\end{split}
\end{equation}
where the parameters $\gamma, \beta \in \mathbb{R}^{E}$ are obtained by a linear projection of a mean-pooled conditioning vector $c_\text{mod} \in \mathbb{R}^{E}$, computed from $c_m$ and $e_t$.

\subsection{Beat Representation} \label{sec:beat_representation}
Prior works have identified a strong consistency between musical beats and the dance motion beats~\cite{fact, bailando, edge, popdg, lodge, beatit}.
Capturing this rhythmic link is essential for coherent and expressive generation.
In many existing approaches~\cite{fact, bailando, edge, popdg, lodge}, beat information is included in the music features.
For instance, Librosa~\cite{librosa} is commonly used to extract music feature $m_\text{total} \in \mathbb{R}^{L \times D_\text{music}}$, where $D_\text{music}=35$ whose last channel dimension corresponds to a binary beat signal $b_\text{raw} \in \mathbb{R}^{L \times 1}$.
Beat-It~\cite{beatit} mitigates the sparsity by encoding, for each frame, the temporal distance to the nearest beat, but the resulting signals are still monotonic and do not explicitly model the decaying influence of off-beat frames.

In practice, beats segment choreography into phrases and anchor high-kinetic movements.
Therefore, an effective representation should satisfy two key properties: (\RNum{1})~\emph{frames closer to beats carry stronger signals, and} (\RNum{2})~\emph{this strength decays rapidly yet smoothly with temporal distance}.
We propose \emph{Gaussian beat representation}, an intuitive beat representation $b_\text{total} \in \mathbb{R}^{L \times 1}$ based on Gaussian decay function, that meets both criteria and yields an interpretable, time-localized cue for decoding.
We omit the subscription ``total'' in this subsection for simplicity.

\begin{figure}[ht]
    \centering
    \includegraphics[width=0.48\textwidth]{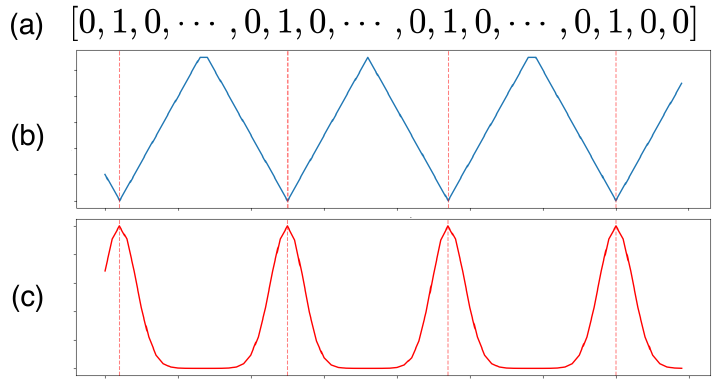}
    \caption{\textbf{Visualizations of raw beat (a), Nearest Beat Distance (NBD) (b), and our Gaussian beat representation (c).} Horizontal axis denotes frame indices (time step) of a sequence and vertical axis indicates signal. The signal range of NBD and the proposed representation are $[0,11]$ and $[0, 1]$, respectively.}
    \label{fig:beat_representation}
\end{figure}

As illustrated in \Fref{fig:beat_representation}, let $b_\text{raw}$ denote the binary beat sequence $b_\text{raw}(i) \in \left\{ 0, 1 \right\}$ at frame $i \in \left\{0, \dots, L-1 \right\}$.
We formulate the Nearest Beat Distance (NBD) in the spirit of Beat-It as the minimum distance from each frame to the closest preceding or following beat frame:
\begin{equation} \label{eq:nbd}
\begin{split}
    &\operatorname{NBD}(i) = \min\!\left( \operatorname{dist}\!^{-}\!(i), \operatorname{dist}\!^{+}\!(i) \right),
\end{split}
\end{equation}
where $\operatorname{dist}\!^{-}\!(i) = i - \max\!\left( j \leq i \vert b_j\!=\!1 \right)$ if previous beat exists else $+\infty$, and $\operatorname{dist}\!^+\!(i) = \min\!\left( j \geq i \vert b_j\!=\!1 \right) - i$ if following beat exists else $+\infty$.
Here, $j$ indexes frames in the sequence.
Let the beat frames be $\tau_0 < \tau_1 < \dots < \tau_{M-1}$.
To cover the full sequence and obtain a local tempo measure, we augment boundaries with $\tau_{-1} = 0$ and $\tau_M = L-1$, and define the inter-beat interval $l(i) = \tau_{k+1} - \tau_k$ for $\tau_k \leq i \leq \tau_{k+1}$.
To satisfy the two desiderata, \textit{i.e.}, stronger signals near beats and smooth, rapidly decaying influence away from beats, while normalizing across different tempos, we use Gaussian decay with smoothing factor $\alpha \in (0, 1)$ for our beat representation:
\begin{equation} \label{eq:beat_representation}
\begin{split}
    b(i) = \operatorname{exp}\left(-\cfrac{\operatorname{NBD}(i)^2}{2 \left(\alpha \cdot l(i)\right)^2}\right).
\end{split}
\end{equation}
The Gaussian provides a bell-shaped smooth, localized emphasis around beat frames, and the tempo-adaptive bandwidth $\alpha \cdot l(i)$ yields consistent guidance under varying beat spacings.
The resulting signal is an explicit and interpretable rhythmic prior that aligns with the phrasing structure of dance motion.

\begin{table*}[!ht]
\centering
\resizebox{0.99\textwidth}{!}{%
\begin{tabular}{l|l|ccc|c|cc|c}
\toprule
\textbf{Dataset} & \textbf{Model} & \multicolumn{3}{c|}{\textbf{Fidelity}} & \textbf{Beat} & \multicolumn{2}{c|}{\textbf{Diversity}} & Wins~($\uparrow$) \\
\cmidrule(lr){3-5} \cmidrule(lr){6-6} \cmidrule(lr){7-8}
& & $\operatorname{FID}_k$ ($\downarrow$) & $\operatorname{FID}_g$ ($\downarrow$) & $\operatorname{PFC}$ ($\downarrow$) & $\operatorname{BAS}$ ($\uparrow$) & $\operatorname{Div}_k$ ($\rightarrow$) & $\operatorname{Div}_g$ ($\rightarrow$) &  \\
\midrule
\multirow{5}{*}{FineDance}
& GT & - & - & 0.1852 & - & 10.9924 & 7.7424 & - \\
\cmidrule(lr){2-9}
& EDGE & 179.01$^{\scriptscriptstyle \pm3.10}$ & 1234.28$^{\scriptscriptstyle \pm236.95}$ & 0.3994$^{\scriptscriptstyle \pm0.0177}$ & 0.2261$^{\scriptscriptstyle \pm0.0013}$ & \textbf{10.34}$^{\scriptscriptstyle \pm0.36}$ & 31.74$^{\scriptscriptstyle \pm2.76}$ & 3.5\% \\
& POPDG & 190.23$^{\scriptscriptstyle \pm1.18}$ & 1479.14$^{\scriptscriptstyle \pm6.44}$ & 0.3765$^{\scriptscriptstyle \pm0.0124}$ & 0.2361$^{\scriptscriptstyle \pm0.0033}$ & 7.22$^{\scriptscriptstyle \pm0.14}$ & 14.53$^{\scriptscriptstyle \pm0.12}$ & 5.5\% \\
& Lodge & \underline{84.99}$^{\scriptscriptstyle \pm2.07}$ & \underline{64.57}$^{\scriptscriptstyle \pm10.74}$ & \underline{0.0585}$^{\scriptscriptstyle \pm0.014}$ & \underline{0.2410}$^{\scriptscriptstyle \pm0.0063}$ & \underline{7.98}$^{\scriptscriptstyle \pm0.20}$ & \textbf{7.67}$^{\scriptscriptstyle \pm0.68}$ & \underline{35.0\%} \\
& \textbf{Ours} & \textbf{51.36}$^{\scriptscriptstyle \pm0.67}$ & \textbf{43.11}$^{\scriptscriptstyle \pm10.54}$ & \textbf{0.0119}$^{\scriptscriptstyle \pm0.0008}$ & \textbf{0.2441}$^{\scriptscriptstyle \pm0.0044}$ & 6.38$^{\scriptscriptstyle \pm0.17}$ & \underline{6.44}$^{\scriptscriptstyle \pm0.88}$ & \textbf{56.0\%} \\
\midrule
\multirow{5}{*}{AIST++}
& GT & - & - & 1.2544 & - & 9.61 & 7.78 & - \\
\cmidrule(lr){2-9}
& EDGE & 125.99$^{\scriptscriptstyle \pm128.69}$ & \underline{28.72}$^{\scriptscriptstyle \pm4.29}$ & 3.1883$^{\scriptscriptstyle \pm0.5318}$ & \underline{0.2572}$^{\scriptscriptstyle \pm0.0112}$ & \textbf{11.45}$^{\scriptscriptstyle \pm3.25}$ & 4.91$^{\scriptscriptstyle \pm0.56}$ & 10.5\% \\
& POPDG & 777.32$^{\scriptscriptstyle \pm711.65}$ & 60.08$^{\scriptscriptstyle \pm5.98}$ & 4.8615$^{\scriptscriptstyle \pm0.6010}$ & 0.2318$^{\scriptscriptstyle \pm0.0129}$ & 24.08$^{\scriptscriptstyle \pm7.40}$ & \textbf{7.87}$^{\scriptscriptstyle \pm0.59}$ & 9.0\% \\
& Lodge & \underline{67.13}$^{\scriptscriptstyle \pm2.79}$ & 28.93$^{\scriptscriptstyle \pm0.47}$ & \underline{1.4087}$^{\scriptscriptstyle \pm0.1296}$ & 0.2397$^{\scriptscriptstyle \pm0.0158}$ & 3.34$^{\scriptscriptstyle \pm0.32}$ & 3.54$^{\scriptscriptstyle \pm0.14}$ & \underline{32.0\%} \\
& \textbf{Ours} & \textbf{65.86}$^{\scriptscriptstyle \pm3.11}$ & \textbf{26.58}$^{\scriptscriptstyle \pm1.02}$ & \textbf{1.0622}$^{\scriptscriptstyle \pm0.2343}$ & \textbf{0.2701}$^{\scriptscriptstyle \pm0.0116}$ & \underline{3.57}$^{\scriptscriptstyle \pm0.37}$ & \underline{4.98}$^{\scriptscriptstyle \pm0.43}$ & \textbf{48.5\%} \\
\bottomrule
\end{tabular}%
}
\caption{\textbf{Quantitative results on the FineDance and AIST++ datasets.} GT motion is used as the reference. $\downarrow$ indicates lower is better, and $\rightarrow$ indicates closer to the real motion is better. The best and second-best results are highlighted in bold and underline, respectively.}
\label{table:all_results}
\end{table*}

\subsection{Training and Inference}
\label{sec:training_and_inference}

\paragraph{Training schemes and losses.}
Following Lodge~\cite{lodge}, we train the global and local diffusion models independently, and use the global key motions (hard cues $d_h$ and soft cues $d_s$) to guide the local model at inference time only.
Directly controlling only a few boundary frames via diffusion inpainting can cause incoherent transitions within each local window, so we fine-tune the local diffusion by replacing the first and last $L_\text{key} / 2$ frames of the noisy input $d_t$ with the corresponding ground-truth frames $d_0$ during training.

We basically use a standard reconstruction loss of the diffusion models~\cite{diffusion} for the dance decoder $f_\theta$ of our \emph{MambaDance}, defined as:
\begin{equation} \label{eq:loss_simple}
\begin{split}
    \mathcal{L}_\text{simple} = \mathbb{E}_{d, t} \left[ \left\Vert d - f_\theta(d_t, t, m, b) \right\Vert_2^2 \right].
\end{split}
\end{equation}
Additional auxiliary losses make training stable and improve physical plausibility, containing position loss $\mathcal{L}_\text{pos}$, velocity loss $\mathcal{L}_\text{vel}$, acceleration loss $\mathcal{L}_\text{acc}$, and contact consistency loss on foot $\mathcal{L}_\text{foot}$.
Position loss measures the similarity of joint positions between ground truth and predicted dance movements:
\begin{align} \label{eq:pos_loss}
\begin{split}
    &\mathcal{L}_\text{pos}\!=\!\cfrac{1}{l} \sum\limits_{i=1}^l \left\Vert \operatorname{FK}(d^{i})\!-\!\operatorname{FK}(\hat{d}^{i}) \right\Vert_2^2,
\end{split}
\end{align}
where $\operatorname{FK}(\cdot)$ denotes the forward kinematic function that converts joint angles into joint positions.
Similarly, the contact consistency loss ensures accurate foot-ground contacts:
\begin{align} \label{eq:foot_loss}
\begin{split}
    &\mathcal{L}_\text{foot}\!=\!\cfrac{1}{\!l\!-\!1\!} \sum\limits_{i=1}^{l-1} \left\Vert (\operatorname{FK}_\text{foot}(\hat{d}^{i+1})\!-\!\operatorname{FK}_\text{foot}(\hat{d}^{i}))\!\cdot\!\hat{y}^{i} \right\Vert_2^2,
\end{split}
\end{align}
where $\operatorname{FK}_\text{foot}(\cdot)$ operates the forward kinematic function for foot joints only, and $\hat{y}$ stands for the predicted binary foot contact label.
The velocity loss and acceleration loss assess the similarity of joint velocities and accelerations:
\begin{align} \label{eq:vel_acc_losses}
\begin{split}
    &\mathcal{L}_\text{vel}\!=\!\cfrac{1}{\!l\!-\!1\!} \sum\limits_{i=1}^{l-1} \left\Vert (d^{i+1}\!-\!d^{i}) - (\hat{d}^{i+1}\!-\!\hat{d}^{i}) \right\Vert_2^2, \\
    &\mathcal{L}_\text{acc}\!=\!\cfrac{1}{\!l\!-\!2\!} \sum\limits_{i=1}^{l-2} \left\Vert (v^{i+1}\!-\!v^{i}) - (\hat{v}^{i+1}\!-\!\hat{v}^{i}) \right\Vert_2^2.
\end{split}
\end{align}
Note that the joint acceleration can be calculated with the joint velocity $v^{i} = d^{i+1} - d^{i}$.
The total loss is defined as combined all losses with weights:
\begin{equation} \label{eq:loss_total}
\begin{split}
    \mathcal{L}_\text{total} = \mathcal{L}_\text{simple} &+ \lambda_\text{pos} \mathcal{L}_\text{pos} + \lambda_\text{vel} \mathcal{L}_\text{vel} \\ &+ \lambda_\text{acc} \mathcal{L}_\text{acc} + \lambda_\text{foot} \mathcal{L}_\text{foot}. 
\end{split}
\end{equation}
To further suppress unnecessary drift in the root, we add a root translation term $\lambda_\text{trans} \mathcal{L}_\text{trans}$ (\Eref{eq:vel_acc_losses} on the root position) when fine-tuning the local model.

\paragraph{Parallel inference with two-stage diffusion model.}
Inspired by two-stage diffusion model for long dance generation~\cite{lodge}, we generalize the inference pipeline to handle variable lengths.
Given total music features $m_\text{total} \in \mathbb{R}^{L \times D_\text{music}}$ and Gaussian beat representations $b_\text{total} \in \mathbb{R}^{L \times 1}$, we partition them into $k$ non-overlapping segments of length $N$.
Recall that the length of the sliced motion and music sequences is $l \in \{N, n\}$, where $N$ for global diffusion and $n$ for local diffusion.
Whereas the prior method targets only long sequences with $(N, n) = (1024, 256)$, our inference supports $N \in \left\{128, 256, 1024\right\}$ and $n \in \left\{64, 256\right\}$, enabling generation at multiple temporal resolutions while preserving the coarse-to-fine design.

Specifically, in the dance decoder, using full mixture of musical condition $c_{m,g} \in \mathbb{R}^{N \times E}$, the global diffusion predicts characteristic key motions $m_\text{key}$ that capture high-level choreographic patterns with elevated kinetic energy.
The key motions $m_\text{key} \in \mathbb{R}^{L_\text{key} \times D_\text{motion}}$ includes hard cues $d_h$ (to anchor window boundaries) and soft cues $d_s$ (to shape intra-window dynamics).
To exploit bilateral symmetry, we mirror $d_s$ and place $d_s$ in a $n$-length sequence for soft guidance.
For sequences spanning multiple segments, we ensure continuity by copying the last $L_\text{key}$ frames of segment $i$ into the first $L_\text{key}$ frames of segment $i+1$ before extracting mid-region primitives.

Each segment $c_{m,g}^i$ is further divided into windows $\{c_{m,l}^j\}_{j=1}^{N/\!/n}$, where $c_{m,l} \in \mathbb{R}^{n\times E}$.
We inject $d_h$ via diffusion inpainting at the start and end of each window to fix boundary poses for reliable stitching, and we apply $d_s$ as early-step guidance for $>\!(1000 \cdot s)$ denoising steps, where soft cue guidance scale $s$ controls the strength.
Because boundaries are anchored, windows decode independently and can be generated in parallel, and the resulting clips are concatenated to form the total dance sequence.

\section{Experiments} \label{sec:4}

\begin{table*}[!ht]
\centering
\resizebox{0.8\textwidth}{!}{%
\begin{tabular}{c|ccc|c|cc}
\toprule
\multirow{2}{*}{\textbf{Model}} & \multicolumn{3}{c|}{\textbf{Fidelity}} & \textbf{Beat} & \multicolumn{2}{c}{\textbf{Diversity}} \\
\cmidrule(lr){2-4} \cmidrule(lr){5-5} \cmidrule(lr){6-7}
& $\operatorname{FID}_k$ ($\downarrow$) & $\operatorname{FID}_g$ ($\downarrow$) & $\operatorname{PFC}$ ($\downarrow$) & $\operatorname{BAS}$ ($\uparrow$) & $\operatorname{Div}_k$ ($\rightarrow$) & $\operatorname{Div}_g$ ($\rightarrow$) \\
\midrule
GT & - & - & 0.1852 & - & 10.9924 & 7.7424 \\
\midrule
Lodge & 84.99$^{\scriptscriptstyle \pm2.07}$ & 64.57$^{\scriptscriptstyle \pm10.74}$ & 0.0585$^{\scriptscriptstyle \pm0.014}$ & 0.2410$^{\scriptscriptstyle \pm0.0063}$ & \textbf{7.98}$^{\scriptscriptstyle \pm0.20}$ & \textbf{7.67}$^{\scriptscriptstyle \pm0.68}$ \\
Mamba-only & \underline{60.94}$^{\scriptscriptstyle \pm2.01}$ & 43.46$^{\scriptscriptstyle \pm7.29}$ & \underline{0.0180}$^{\scriptscriptstyle \pm0.0017}$ & 0.2402$^{\scriptscriptstyle \pm0.0044}$ & \underline{6.48}$^{\scriptscriptstyle \pm0.40}$ & \underline{7.21}$^{\scriptscriptstyle \pm0.79}$ \\
Mamba+NBD & 83.55$^{\scriptscriptstyle \pm1.35}$ & \textbf{38.43}$^{\scriptscriptstyle \pm0.71}$ & 0.0687$^{\scriptscriptstyle \pm0.0046}$ & \textbf{0.2478}$^{\scriptscriptstyle \pm0.0054}$ & 5.3958$^{\scriptscriptstyle \pm0.17}$ & 5.78$^{\scriptscriptstyle \pm0.30}$ \\
\textbf{Ours} & \textbf{51.36}$^{\scriptscriptstyle \pm0.67}$ & \underline{43.11}$^{\scriptscriptstyle \pm10.54}$ & \textbf{0.0119}$^{\scriptscriptstyle \pm0.0008}$ & \underline{0.2441}$^{\scriptscriptstyle \pm0.0044}$ & 6.38$^{\scriptscriptstyle \pm0.17}$ & 6.44$^{\scriptscriptstyle \pm0.88}$ \\
\bottomrule
\end{tabular}%
}
\caption{\textbf{Ablation on Mamba and beat representation (FineDance).} In contrast to the Transformer-based Lodge~\cite{lodge}, ``Mamba-only'' denotes a fully Mamba-based model without an explicit beat prior. ``Mamba+NBD'' uses Nearest Beat Distance~\cite{beatit} as a beat representation.}
\label{table:ablation_mamba_and_beat}
\end{table*}

\subsection{Experiment Design}
We evaluate \emph{MambaDance} on AIST++~\cite{fact} and FineDance~\cite{finedance}.
AIST++ consists of 1{,}408 high-quality short 3D dance sequences, performed by professional dancers across 10 genres.
We use a sliced motion parameterization of $D_{\text{motion}}=151$ following EDGE~\cite{edge} and train with sequence length $N=128$.
FineDance provides long-form dance sequences collected from online videos, covering diverse dance styles and music durations, spanning 16 genres.
We use 139-dimensional representation with sequence length $N=1024$ as in Lodge~\cite{lodge}.
These datasets jointly cover complementary regimes--short and long clips at 30~FPS--allowing us to assess robustness across temporal resolutions and data distribution.

We compare against recent Transformer-based diffusion approaches for music-conditioned 3D dance generation:our method with the following baselines, which show recent advances by leveraging Transformer-based diffusion architectures in music-conditioned 3D dance generation.
\begin{itemize}

    \item \textbf{EDGE}~\cite{edge}: The first approach to use Transformer-based diffusion model for 3D dance generation with rich music representation from Jukebox.

    \item \textbf{POPDG}~\cite{popdg}: A follow-up method utilizing an improved diffusion model (iDDPM~\cite{iddpm}) based on additional alignment module and space augmentation algorithm.

    \item \textbf{Lodge}~\cite{lodge}: An EDGE-based two-stage diffusion framework with foot refine block and a multi genre discriminator, focusing on long-term dance generation.
\end{itemize}


\subsection{Evaluation Metrics}
We evaluate generated dances using standard metrics following EDGE and Lodge.
To measure motion realism, we report Fréchet Inception Distance, computed over kinematic ($\operatorname{FID}_k$) and geometric ($\operatorname{FID}_g$) variants.
Physical plausibility is assessed using the Physical Foot Contact score ($\operatorname{PFC}$).
The Beat Alignment Score ($\operatorname{BAS}$) measures how well the generated motion beats--moments where motion energy peaks (e.g., local maxima of joint speed or kinetic energy)--align with the music beats.
Motion diversity is quantified by the Diversity metric, which computes the average pairwise distance over kinematic ($\operatorname{Div}_k$) and geometric ($\operatorname{Div}_g$) features.
Since the metrics do not exactly reflect human perception, we conduct user studies and report the win rate ($\operatorname{Wins}$) of ours over the baselines.

Unlike prior works~\cite{edge, popdg, lodge}, we calculate all metrics on full-length sequences rather than sliced with length $N$, better reflecting global temporal coherence, aligned with the goal of generating complete dances for an entire music.
In addition, we report mean and standard deviation across 10 independent runs to indicate statistical reliability.
Further details regarding the evaluation metrics and the user study protocol are provided in the supplementary material.


\begin{figure}[ht]
    \centering
    \includegraphics[width=\linewidth]{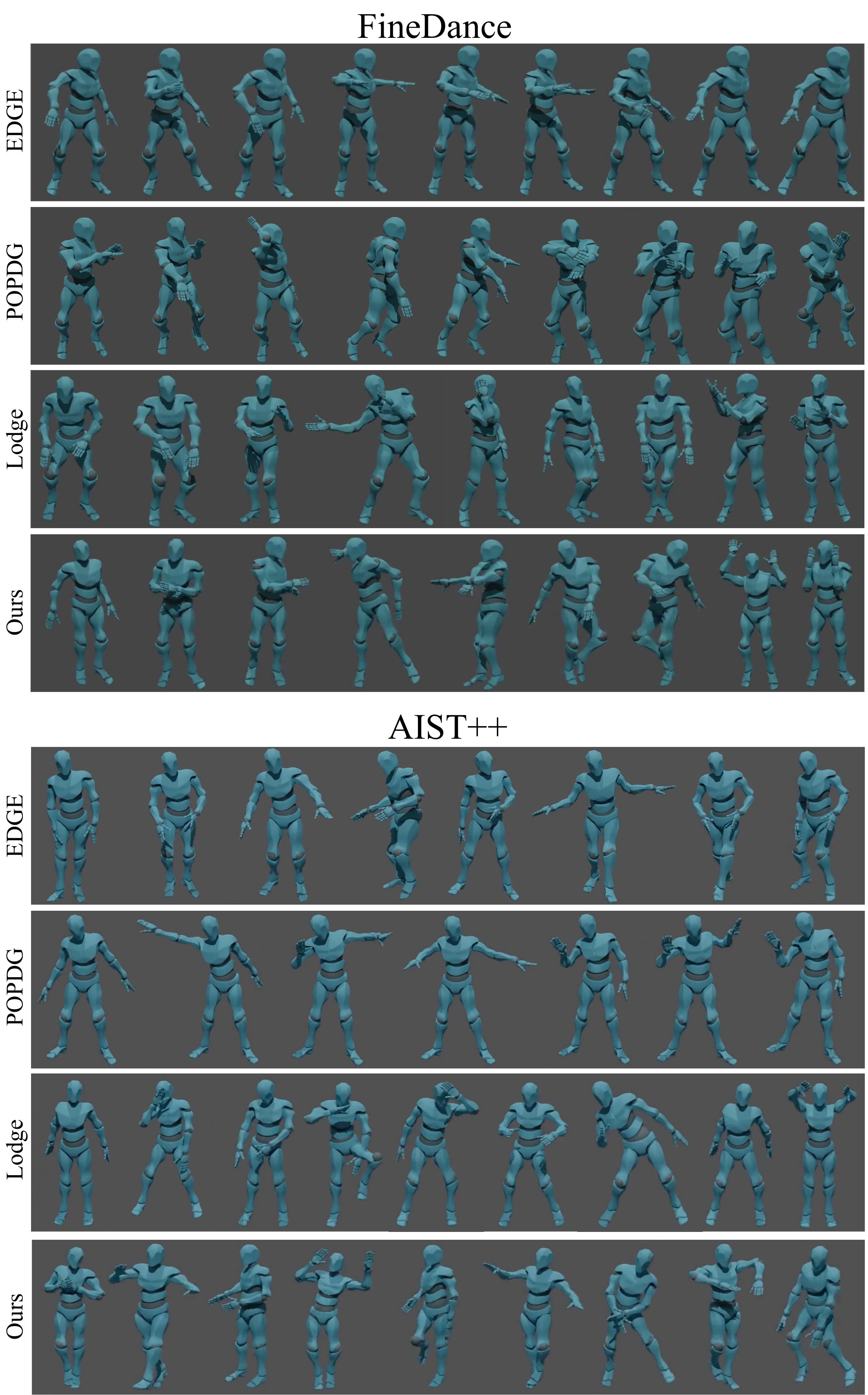}
    \caption{\textbf{Qualitative comparison on the FineDance (top) and AIST++ (bottom) dataset.} 
    Each row shows a set of sampled frames captured at consistent intervals from the full sequence.}
    \label{fig:qualitative_comparison}
\vspace{-5mm}
\end{figure}

\begin{table*}[!ht]
\resizebox{0.75\linewidth}{!}{%
\begin{tabular}{cc|ccccc}
\toprule
\multicolumn{2}{c|}{\textbf{Ablations}} & \multicolumn{5}{c}{\textbf{Metrics}} \\[0.8ex]
AdaLM & CMM & $\operatorname{FID}_k$ ($\downarrow$) & $\operatorname{PFC}$ ($\downarrow$) & $\operatorname{BAS}$ ($\uparrow$) & $\operatorname{Div}_k$ ($\rightarrow$) & $\operatorname{Div}_g$ ($\rightarrow$) \\
\midrule
\multicolumn{2}{c|}{GT} & - & - & 0.1852 & 10.99 & 7.74 \\
\midrule
\xmark & \xmark & \underline{61.28}$^{\scriptscriptstyle \pm1.72}$ & \underline{0.6316}$^{\scriptscriptstyle \pm0.0066}$ & \underline{0.2465}$^{\scriptscriptstyle \pm0.0044}$ & \underline{8.59}$^{\scriptscriptstyle \pm0.16}$ & 5.77$^{\scriptscriptstyle \pm0.41}$ \\
\xmark & \cmark & 62.93$^{\scriptscriptstyle \pm1.80}$ & 0.7878$^{\scriptscriptstyle \pm0.0070}$ & \textbf{0.2479}$^{\scriptscriptstyle \pm0.0057}$ & \textbf{9.93}$^{\scriptscriptstyle \pm0.18}$ & \underline{6.11}$^{\scriptscriptstyle \pm0.60}$ \\
\cmark & \cmark & \textbf{51.36}$^{\scriptscriptstyle \pm0.67}$ & \textbf{0.0119}$^{\scriptscriptstyle \pm0.0008}$ & 0.2441$^{\scriptscriptstyle \pm0.0044}$ & 6.38$^{\scriptscriptstyle \pm0.17}$ & \textbf{6.44}$^{\scriptscriptstyle \pm0.88}$ \\
\bottomrule
\end{tabular}
}
\centering
\caption{\textbf{Ablation study on different decoder structures.} The last row corresponds to our \emph{MambaDance}.}
\label{table:ablation_modules}
\end{table*}

\subsection{Comparisons}
We evaluate \emph{MambaDance} with recent Transformer-based diffusion models, EDGE, POPDG, and Lodge, on AIST++ and FineDance.
The global/local lengths are $(N, n) = (128,64)$ for AIST++ and $(1024,256)$ for FineDance.

Across both datasets, our method achieves the best scores on motion fidelity and beat alignment ($\operatorname{FID_k}$, $\operatorname{FID}_g$, $\operatorname{PFC}$, $\operatorname{BAS}$) (\Tref{table:all_results}).
On FineDance, our model reduces $\operatorname{FID}_k$ to 51.36 (vs.\ 84.99 for Lodge) and $\operatorname{FID}_g$ to 43.11 (vs.\ 64.57), and lowers $\operatorname{PFC}$ to 0.0119 (vs.\ 0.0585), indicating more realistic dynamics and substantially more stable foot-ground interaction.
$\operatorname{BAS}$ increases slightly to 0.2441 (vs.\ 0.2410 for Lodge), reflecting a consistent and noticeable gain in rhythm-motion alignment.
On AIST++, \emph{MambaDance} attains 65.86 $\operatorname{FID}_k$ (vs.\ 67.13 for Lodge), 26.58 $\operatorname{FID}_g$ (vs.\ 28.72 for EDGE), 1.0622 $\operatorname{PFC}$ (vs.\ 1.4087 for Lodge), and the highest $\operatorname{BAS}$ at 0.2701 (vs. 0.2572 for EDGE).
Notably, our standard deviations are small across metrics, suggesting stable behavior over runs, whereas some baselines show very large variance, e.g., $\operatorname{FID}_k$ on AIST++ and $\operatorname{FID}_g$ on FineDance for EDGE and POPDG.

In terms of the diversity, $\operatorname{Div}_g$ 7.67 of Lodge is closest to GT (7.74), with ours second (6.44) on FineDance.
On AIST++, $\operatorname{Div}_g$ 7.87 of POPDG is closest to GT (7.78), and ours is second (4.98).
EDGE and POPDG often reports the highest diversity, but these cases coincide with elevated $\operatorname{FID}$/$\operatorname{PFC}$, indicating that some of the extra variability stem from artifacts such as foot sliding.
Overall, our method favors physical plausibility and beat alignment, with diversity values that are competitive but somewhat conservative relative to GT.

Results support two conclusions: (\RNum{1}) replacing attention with state-space decoding improves autoregressive motion fidelity and physical plausibility according to the large improvements in $\operatorname{PFC}$; and (\RNum{2}) an explicit beat prior yields consistently large $\operatorname{BAS}$ gains.
The diversity is balanced rather than maximized, trading off against fidelity and physics.
Furthermore, as reflected in the $\operatorname{Wins}$ metric from our user study, dance sequences generated by our model are consistently preferred by human evaluators over those from baseline methods. Notably, our approach maintains stable performance across both short and long sequences, whereas EDGE and POPDG tend to degrade in quality when generating longer videos.

As described in \Fref{fig:qualitative_comparison}, although diversity metrics may suggest reduced variation, the qualitative results reveal dynamic and rhythm-aligned movements.
EDGE and POPDG frequently exhibit artifacts such as foot-sliding that coincide with elevated $\operatorname{FID}$ and $\operatorname{Div}$ metrics and unnaturally staring at one side, as visible in the supplementary videos.
{We strongly encourage watching the supplementary videos for complete and detailed comparisons.}

\subsection{Ablation Studies}
We analyze the effects of the proposed beat representation and model architecture on the FineDance dataset.

We use three variants on FineDance~\cite{finedance}: (\RNum{1}) Lodge~\cite{lodge}, a Transformer-based two-stage diffusion model, (\RNum{2}) ``Mamba-only'' which fully replaces Transformer~\cite{transformer} to Mamba~\cite{mamba, mamba2} without any beat representation, and (\RNum{3}) ``Mamba+NBD'' which uses Nearest Beat Distance~\cite{beatit} in place of the Gaussian beat representation with our Music-Beat Mixer.
Compared to Lodge, ``Mamba-only'' markedly improves motion fidelity and physics (lower $\operatorname{FID}_k$, $\operatorname{FID}_g$, $\operatorname{PFC}$) with only a minor change in $\operatorname{BAS}$, indicating that state-space model makes a dance decoder generate more realistic dynamics and more stable foot-ground interaction.
Introducing the NBD prior yields the highest $\operatorname{BAS}$ and the lowest $\operatorname{FID}_g$, but it degrades not only kinematic realism and stability a lot (e.g., even higher $\operatorname{PFC}$ than Lodge) but also the diversity (farther $\operatorname{Div}_k$ and $\operatorname{Div}_g$ from GT).
Our Gaussian beat representation strikes a balance: $\operatorname{BAS}$ remains high (second-best and close to NBD), while $\operatorname{FID}_k$ and $\operatorname{PFC}$ improve to the best values and diversity increases.
Taken together, these results suggest that Mamba is the primary driver of fidelity gains, and our Gaussian beat prior preserves rhythm alignment without sacrificing naturalness.

We ablate the Cross-Modal Mamba (CMM) which is our replacement of cross-attention and Adaptive Linear Modulation (AdaLM) which is our group normalization based linear modulation.
When a model is designed with CMM instead of cross-attention module, $\operatorname{BAS}$ and diversity raise (more varied motion and tighter audio-motion coupling).
Adding AdaLM on top of CMM recovers stability and yields the best overall fidelity and plausibility, with BAS remaining comparable and diversity decreasing modestly.
This pattern indicates that CMM supplies effective music-motion fusion, while AdaLM regularizes the fused states, acting as a lightweight stabilizer for contact and dynamics.

\section{Conclusion}
\label{sec:5}

In this paper, we have proposed \emph{MambaDance}, a novel approach for music-to-3D dance generation, fully substituting Mamba for Transformer.
Furthermore, we introduce an informative beat representation based on Gaussian decay, considering the important nature of musical beats.
Experimental results on two datasets with different sequence lengths demonstrate the robustness and superiority of our approach over baselines.
Although there is potential for advancements in applications, our work focuses on generating human motion conditioned on music, not addressing the downstream stages of the production pipeline, such as rendering.
A natural extension can be an end-to-end system that maps music to a dancing 3D avatar, including a rendering stack for camera, materials, and lighting.

\section*{Acknowledgements}
\label{sec:6}

This work was supported by Institute of Information \& communications Technology Planning \& Evaluation (IITP) grant funded by the Korea government (MSIT) (No.RS-2020-II201336, Artificial Intelligence Graduate School Program (UNIST); No.RS-2022-II220612, Geometric and Physical Commonsense Reasoning based Behavior Intelligence for Embodied AI; No.RS-2025-25442149, LG AI STAR Talent Development Program for Leading Large-Scale Generative AI Models in the Physical AI Domain; No.RS-2025-25442824, AI Star Fellowship Program (UNIST)), and by the InnoCORE program of the Ministry of Science and ICT (25-InnoCORE-01).

\clearpage
\setcounter{page}{1}
\maketitlesupplementary

\section{Preliminaries} \label{sec:preliminaries}
\paragraph{Selective State Space Model.}
State Space Models (SSMs), particularly Structured State Space Models (S4~\cite{S4}) and Mamba~\cite{mamba, mamba2}, have shown superior capabilities of modeling long-range dependencies of sequential data.
These models map an input sequence $x_t \in \mathbb{R}^{T}$ to an transited output sequence $y_t \in \mathbb{R}^T$ through a hidden state $h_t \in \mathbb{R}^{N}$.
SSM can be discretized with step size $\Delta$ as follows:
\begin{equation} \label{eq:ssm}
\begin{split}
    h_t &= A h_{t-1} + B x_t \\ y_t &= C^\top h_t,
\end{split}
\end{equation}
where $A \in \mathbb{R}^{N \times N}$, $B \in \mathbb{R}^{N \times 1}$, and $C \in \mathbb{R}^{N \times 1}$ are state matrix, input matrix, and output matrix, defined by state dimension $N$, respectively.
This system can be expressed using a global convolution with a structured convolutional kernel $\bar{K}$ (note that $x$ denotes general sequential input here):
\begin{equation} \label{eq:ssm_conv}
\begin{split}
    \bar{K} &= (C^\top \bar{B}, C^\top \bar{A} \bar{B}, \dots, C \bar{A}^{L-1} \bar{B}) \\ y &= x * \bar{K}.
\end{split}
\end{equation}

To deviate from linear time-invariance (LTI), Mamba1~\cite{mamba} introduces selective scanning with time-varying parameters, overcoming computational challenges with associative scans.
Mamba2~\cite{mamba2} further enhances the efficiency by conceptually connecting SSM and attention mechanism, enabling faster computations while maintaining competitive performance against Transformers~\cite{transformer}.

\paragraph{Diffusion Model.}
We adopt DDPM~\cite{diffusion} formulation, defined by a forward noising process of latents $\{z_t\}_{t=1}^T$:
\begin{equation} \label{eq:diffusion_forward_process}
\begin{split}
    q(z_t \vert x) \sim \mathcal{N}(\sqrt{\bar{\alpha}_t} x, (1 - \bar{\alpha}_t) \mathbf{I}),
\end{split}
\end{equation}
where $x \sim p(x)$, and $\bar{\alpha}_t \in (0, 1)$ are constants which follow a monotonically decreasing schedule.
Given musical condition $c_m$ from music feature $m$ and beat representation $b$, the diffusion model reverses the forward diffusion process to estimate $\hat{x}_\theta(z_t, t, m, b) \approx x$ for all timestep $t$, where $\theta$ denotes the model parameters.

We adopt a standard reconstruction loss of the diffusion models, defined as:
\begin{equation} \label{eq:loss_simple}
\begin{split}
    \mathcal{L}_\text{simple} = \mathbb{E}_{x, t} \left[ \left\Vert x - \hat{x}_\theta(z_t, t, m, b) \right\Vert_2^2 \right].
\end{split}
\end{equation}

\section{Implementation Details} \label{sec:implementation_details}

We report hyperparameters as $(\operatorname{FineDance}, \operatorname{AIST\small{+}\small{+}})$.
The global sequence length is $N = (1024, 128)$, and the local window size is $n = (256, 64)$; both music and motion are processed at 30FPS.
The global diffusion stage outputs 13 key motions $m_\text{key}$ per sequence--5 hard ques $d_h$ and 8 soft ques $d_s$.
The key-motion length is $L_\text{key} = (8, 4)$, chosen relative to $N$.
After choreographic augmentation, each $d_s$ is mirrored to produce 16 soft-cue instances and placed at beat-aligned locations.
Both global and local models are optimized with Adan~\cite{adan} at learning rate of $2 \times 10^{-4}$.
At inference, we use DDIM~\cite{ddim} sampling with 50 number of inference steps.
The loss weights of each terms are as: $\lambda_\text{pos} = (1, 0.636)$, $\lambda_\text{vel} = (2.964, 2.964)$, $\lambda_\text{acc} = (2.964, 2.964)$, $\lambda_\text{foot} = (20, 10.942)$, and $\lambda_\text{trans} = (0.5, 0.5)$.

On the other hand, two diffusion models have same structure of dance decoder, except for the detail of SMM.
We omit the Spatial SSM block of global diffusion, because the input sequence length and output sequence length are different (recall that Spatial SMM block processes $a' \in \mathbb{R}^{l \times E}$).

\section{Evaluation Metrics} \label{sec:metrics}

To quantitatively evaluate the quality of the generated dance motions, we adopt several commonly used metrics from prior works.
We used a sequence length of 128, which slightly differs from the original baseline setting of 150, and calculated all metrics for whole integrated dance, so the metric values may differ from those reported in prior works.

\paragraph{Motion Quality.} To evaluate the quality of generated motions, we compute the Fréchet Inception Distance ($\operatorname{FID}$) between motion features of generated and ground truth motion sequences. For each motion, we extract kinematic and geometric features, which respectively capture physical naturalness and overall dance choreography. 
\paragraph{Physical Foot Contact Score.} To evaluate the physical plausibility of foot movements in response to dance motion, we adopt the Physical Foot Contact Score ($\operatorname{PFC}$) proposed in EDGE~\cite{edge}. This physically-inspired metric assesses whether foot-ground interactions are realistic or not without requiring explicit physical modeling. It evaluates the center of mass (COM) acceleartion along both horizontal plane and vertical axiz. Lower $\operatorname{PFC}$ scores indicate more physically plausible motions.
\paragraph{Physical Body Contact Score.} Inspired by POPDG~\cite{popdg}, $\operatorname{PBC}$ measures the overall physical feasibility of full-body movements by analyzing inter-limb and upper-body contacts to identify implausible interpenetrations or unnatural poses.
\paragraph{Motion Diversity.} To assess the diversity of the generated motions, we compute the average feature distance of generated motions and ground truth motions. Following Bailando~\cite{bailando}, we consider both kinematic and geometric features, denoted as $\operatorname{Div}_k$ and $\operatorname{Div}_g$, repectively. Higher values indicate greater variability in motion patterns.
\paragraph{Beat Alignment Score.} To evaluate the beat consistency between the generated dance and the music, we follow Bailando~\cite{bailando} and compute the average temporal distance between each music beat and its nearest motion beat. A higher $\operatorname{BAS}$ value indicates better synchronization between the motion and the rhythm of the music.

\paragraph{User Study (Wins)} 
For the user study, we gathered 20 participants.
Each participant evaluated dance videos generated from 2 datasets, 10 music tracks, and 4 models (\emph{MambaDance}, EDGE~\cite{edge}, POPDG~\cite{popdg}, and Lodge~\cite{lodge}).
In total, every participant watched 
$2 \times 10 \times 4$ dance videos, where each set consisted of four sequences generated for the same music by the four models.

Participants were asked to select the best video in each set according to the following criteria:

\begin{itemize}
\item Which video demonstrates the most natural dance movements?
\item Which video aligns best with the music in terms of beat and rhythm synchronization?
\item Which video exhibits the most diverse and dynamic movements?
\end{itemize}
To mitigate positional bias, the order of the four videos within each set was randomized.


{
    \small
    \bibliographystyle{ieeenat_fullname}
    \bibliography{main}
}

\end{document}